\def\year{2019}\relax
\documentclass[letterpaper]{article} 
\usepackage{aaai19}  
\usepackage{times}  
\usepackage{helvet}  
\usepackage{courier}  
\usepackage{url}  
\usepackage{graphicx}  
\usepackage{multirow}
\usepackage{amsmath}
\usepackage{amssymb}
\usepackage[ruled]{algorithm2e}
\usepackage[colorlinks,linkcolor=blue]{hyperref}
\DeclareMathOperator*{\argmax}{argmax}
\newcommand{\eg}{\emph{e.g. }}
\newcommand{\ie}{\emph{i.e. }}
\newcommand{\etal}{\emph{et al.}}
\begin{document}
%
\title{Horizontal Pyramid Matching for Person Re-identification}
\author{Yang Fu$^{1}$, Yunchao Wei$^{1}$\thanks{corresponding author}, Yuqian Zhou$^1$, Honghui Shi$^{1}$ \\ Gao Huang$^3$, Xinchao Wang$^4$, Zhiqiang Yao$^5$, Thomas Huang$^1$ \\ 
{\small $^1$IFP, Beckman, UIUC, $^2$Cornell University, $^3$Stevens Institute of Technology, $^4$CloudWalk Technology}\\
{\tt\small \{yangfu2, yunchao, yuqian2, t-huang1\}@illinois.edu, shihonghui3@gmail.com} \\
{\tt \small gh349@cornell.edu, xinchao.wang@stevens.edu, yaozhiqiang@cloudwalk.cn}
}
\maketitle

\begin{abstract}
\noindent Despite the remarkable recent progress, person re-identification (Re-ID) approaches are still suffering from the failure cases where the discriminative body parts are missing. To mitigate such cases, we propose a simple yet effective Horizontal Pyramid Matching (HPM) approach to fully exploit various partial information of a given person, so that correct person candidates can be still identified even even some key parts are missing. Within the HPM, we make the following contributions to produce a more robust feature representation for the Re-ID task: 1) we learn to classify using partial feature representations at different horizontal pyramid scales, which successfully enhance the discriminative capabilities of various person parts; 2) we exploit average and max pooling strategies to account for person-specific discriminative information in a global-local manner. To validate the effectiveness of the proposed HPM, extensive experiments are conducted on three popular benchmarks, including Market-1501, DukeMTMC-ReID and CUHK03. In particular, we achieve mAP scores of 83.1\%, 74.5\% and 59.7\% on these benchmarks, which are the new state-of-the-arts. Our code is available on \href{https://github.com/OasisYang/HPM}{Github} .

\end{abstract}
\section{Introduction}
\begin{figure}[t]
	\centering
	\includegraphics[width=0.48\textwidth]{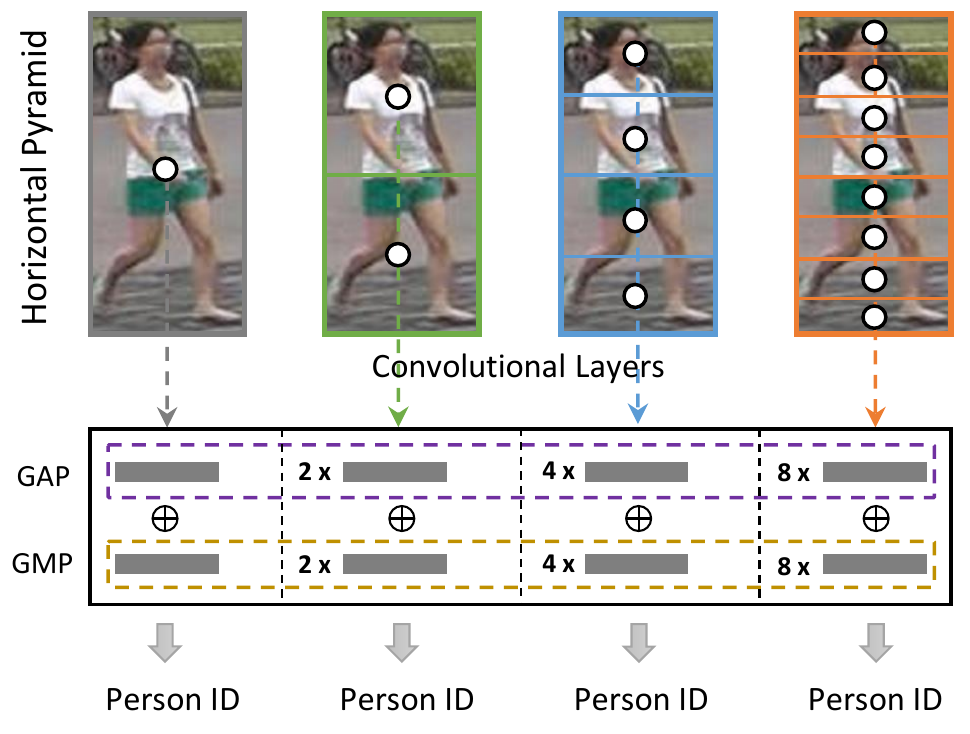}
	\caption{Illustration of the proposed Horizontal Pyramid Matching. We split a {person} into different horizontal parts of multiple scales. The feature representations produced by Global Average Pooling (GAP) and Global Max Pooling (GMP) of each part are then leveraged to learn to the person Re-ID independently.}
	\vspace{-5mm}
	\label{fig:moti}
\end{figure}

Person re-identification (Re-ID) aims at re-identifying a query person from a collection of images, which are taken by multiple cameras across time. It is challenging to learning robust feature representations~\cite{yang2009linear} for each person because of large variations of human attributes like poses, gaits, clothes, as well as the environmental settings like illumination, complex background, and occlusions. 
To address the complexities of visual cues, deep-based approaches~\cite{ahmed2015improved,hermans2017defense,ding2015deep,chen2017multi,li2014deepreid} provide promising solutions. However, these approaches only take advantage of global person features, which turns out to be sensitive to the missing key parts. 


To relieve such issues, many recent approaches have been focusing on learning partial discriminative feature representations. These methods usually take advantage of both global features like body size and local ones like cloths logo, to enhance the robustness of the Re-ID methods. They can be categorized into three types based on the local-region generation scheme. In the first type, prior knowledge like poses or body landmarks are estimated and extracted to localize the discriminative regions~\cite{su2017pose,zheng2017pose,wei2017glad}. However, the performance of Re-ID in this case highly relies on the robustness of the pose or landmark estimation models. Unexpected errors like erroneous estimation of poses may greatly influence the identification result. The second type, attention-based approaches ~\cite{liu2016end,liu2017hydraplus,Zhao2017DeeplyLearnedPR,li2018harmonious,fu2019sta}, focus on extracting the salient regions of interest (ROI) adaptively by localizing the high activations in the deep feature maps. The selected regions however lack semantic interpretation. The third type of methods crop deep feature maps into pre-defined patches or stripes by assuming the images are perfectly aligned~\cite{sun2017beyond,li2017learning}, and are thus prone to errors introduced by outliers.

\begin{figure*}[t]
	\centering
	\includegraphics[width=0.75\textwidth]{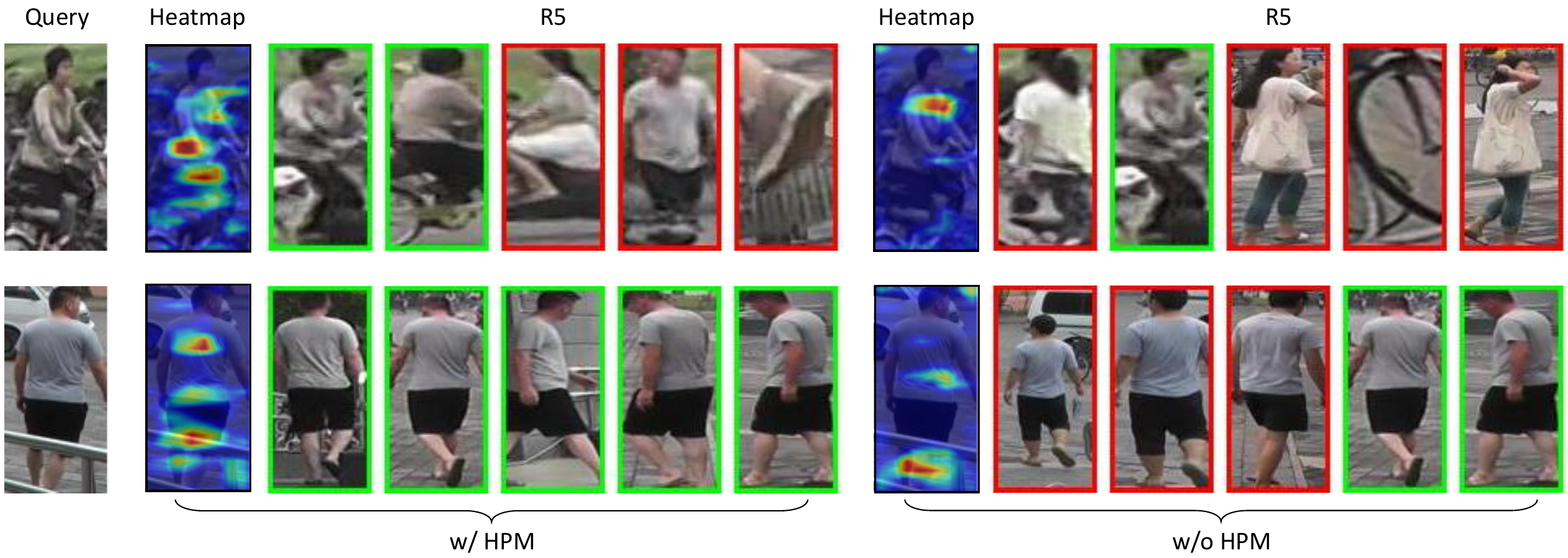}
	\caption{Comparisons of results w/ HPM and w/o HPM in Person Re-ID.}
	\vspace{-5mm}
	\label{fig:demos}
\end{figure*}

To effectively learn partial discriminative features and eliminate the negative effect caused by unexpected pose-variant and unaligned cases, we propose a simple yet effective approach, called Horizontal Pyramid Matching (HPM). Our HPM aim to simultaneously exploit global and partial information of a person for the Re-ID task in a more robust and efficient manner. Specifically, we make three contributions as follows:


\begin{itemize}
\item We horizontally slice the deep feature maps into multiple spatial bins using various pyramid scales for the following pooling operation, which is named as Horizontal Pyramid Pooling (HPP), and learn to classify each spatial bins features output from different pyramid scales independently. Intuitively, using multiple scales of bins will incorporate a slack distance to mitigate the outliers issue caused by misalignment. Also, learning multi-scale information independently will enhance the discriminative information learned in all the scale-specific person parts.

\item We combine the average and max pooling features in each partition. In particular, average pooling is able to perceive the global information of each spatial bin and takes the background context into account. In contrast, the max pooling targets on extracts the most discriminative information and ignore those interference information, mainly coming from similar clothing or background. Integrating them both thus balance the effectiveness of these two strategies in a global-local manner.

\item We evaluate our proposed method on three mainstream person re-identification datasets, Market1501, DukeMTMC-ReID and CUHK03(with new protocol).The experimental results show that our model
beats most of state-of-arts approaches by end-to-end trainable.
\end{itemize}

We illustrate our HPM with one example shown in Figure~\ref{fig:moti}. We first extract the feature representation of the given image with multiple convolutional layers and horizontally slice the feature maps at different pyramid scales. The feature representations generated by both global average pooling and max pooling of each partial scrip are then employed to conduct Re-ID independently. By learning the HPM in such a manner, the partial discriminative capability can be enhanced in a more effective way, which can overcome the disadvantages (\eg sensitive to the missing key parts or misalignment) of current solutions . Figure~\ref{fig:demos} shows the heatmaps of the last convolutional feature maps learned by w/ HPM and w/o HPM schemes. It is observed that more discriminative parts can be identified by our HPM, which leads to better person Re-ID results. 

Extensive experiments and ablation study conducted on Market-1501, DukeMTMC-ReID and CUHK03 have demonstrated the effectiveness of each design. In particular, we achieve the mAP scores of 83.1\%, 74.5\% and 59.7\% on the three benchmarks, which outperform the state-of-the-arts more than 1.5\%, 5.3\% and 2.2\%, respectively.

\section{Related Work}
In this section, we review several closely related work including deep learning methods for Person Re-ID, part-based models, spatial pyramid pooling.

\subsection{Deep learning for Person Re-ID}
Deep learning based method has dominated in Re-ID community \cite{zheng2016person}. Yi~\cite{yi2014deep} first employed deep neural network to determine if a pair of input images belong to the same ID. In general, two types of models are used for person re-identification: verification and identification model. 

For the verification model, they adopt siamese neural network or triplet loss to pull the pair of images with same identity and push away that with different identity~\cite{ahmed2015improved,hermans2017defense,ding2015deep,chen2017multi,li2014deepreid}. In~\cite{hermans2017defense}, Hermans~\etal. proposed a variant of triplet loss to perform end-to-end deep metric learning, which can outperforms many other published methods by a large margin. However, generally, this kind of model may have a compromised efficiency on large gallery. This is because it does not make full use of Re-ID annotations. 

For the identification model~\cite{xiao2016learning,zheng2017discriminatively,sun2017beyond}, it tries to learn a discriminative representation of given input image and it always yields superior accuracy compared with verification model. Xiao~\etal~\cite{xiao2016learning} propose a novel dropout strategy to train a classification model with multiple datasets jointly. In~\cite{zheng2017discriminatively}, the verification and classification losses are combined together to learn a discriminative embedding and a similarity measurement at the same time. In~\cite{sun2017beyond}, a Part-based Convolutional network is proposed to learn discriminative part-informed features.

\subsection{Part-based Model}
Recently, many works generate deep representation from local parts for fine-grained discriminative features of person. This kind of part-based model can be divided into three categories. First one is based on some prior knowledge like pose estimation and landmark detection~\cite{zheng2017pose,su2017pose,wei2017glad}. These methods share a same drawback that is the gap lying between datasets for pose estimation and person retrieval. Second, several other works abandon the semantic cues for partition~\cite{yao2017deep,liu2017hydraplus,Zhao2017DeeplyLearnedPR,li2018harmonious}. For example,  Yao~\etal~\cite{yao2017deep} employed the Part Loss Networks which enforces the deep network to learn representations for different parts and gain the discriminative power on unseen persons. Third, the partition is cropped into pre-defined patches~\cite{sun2017beyond,li2017learning}. Sun\etal~\cite{sun2017beyond} proposed Part-based Convolutional Baseline (PCB) to learn  discriminative partition features. However, the PCB may suffer some outliers, which make the inconsistency in each partition, thus they proposed Refined Part Pooling (RPP) to enhance within-part consistency.

\subsection{Spatial Pyramid Pooling}
Since convolutional neural networks with the fully connectedly layer always require the fixed input size. In order to remove this constrain, He~\etal~\cite{he2014spatial} proposed the Spatial Pyramid Pooling network, which is able to generate a fixed length output regardless of the input size and maintain spatial information by pooling in local spatial bins. Multi-level spatial pooling has also shown to be robust to object deformations. It can improve the performance of classification and object detection tasks. Similarly pyramid pooling module is also used in~\cite{zhao2017pyramid}, the pyramid level pooling separates the feature map into different sub-regions and forms pooled representation for different locations.

\begin{figure*}[t]
	\centering
	\includegraphics[width=1\textwidth]{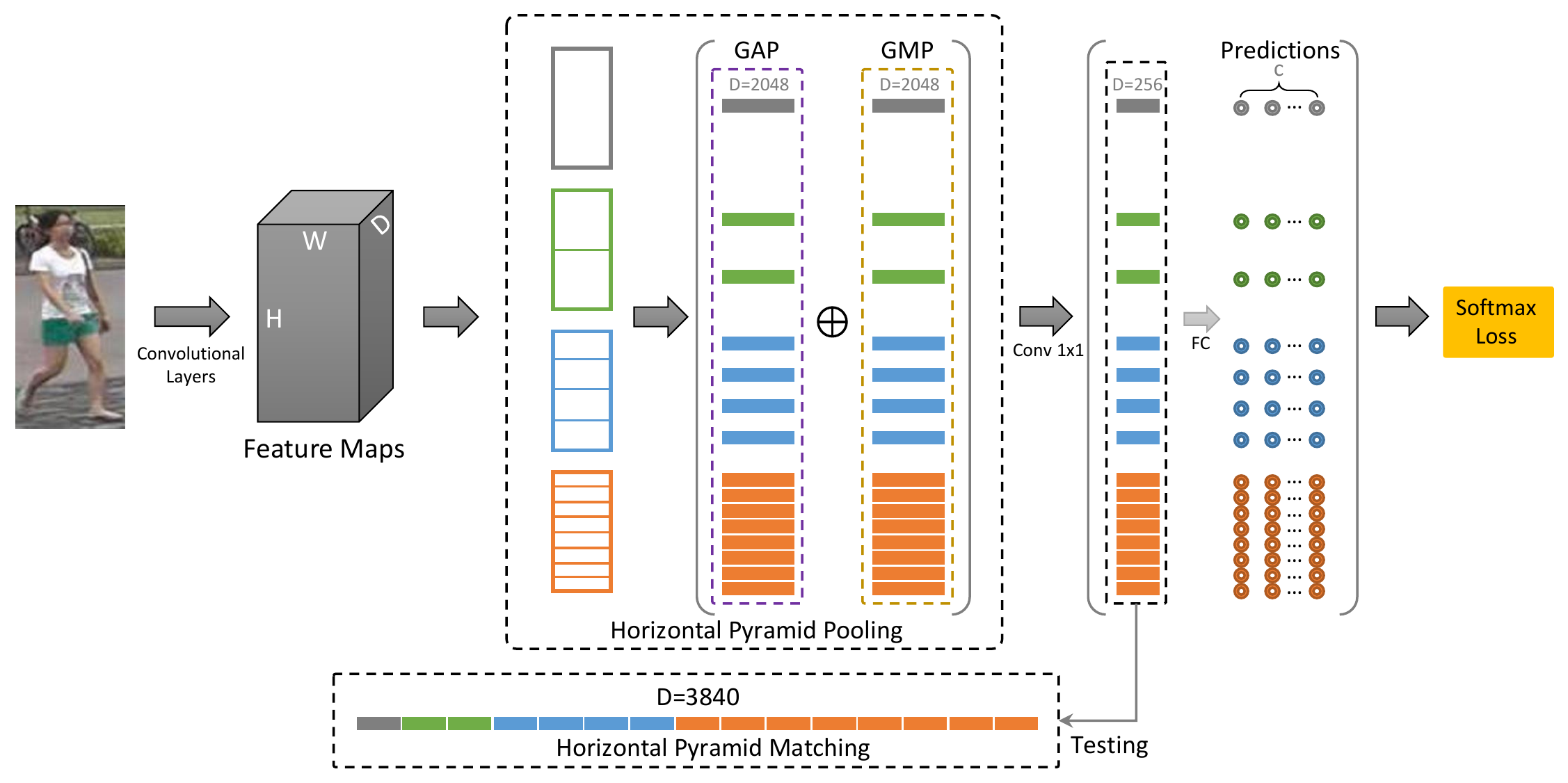}
	\caption{Overview of the proposed Horizontal Pyramid Matching (HPM) approach. The input image firstly goes through a convolutional neural network to extract its feature maps. Then, the horizontal pyramid pooling is leveraged to producing feature representation of each part using both global average pooling and global max pooling. Finally, prediction of each part is fed into the classifier to conduct partial-level person Re-ID. During the testing stage, we concatenate features of parts at different pyramid scales to form the final feature representation of each image.}
	\label{fig:framework}
	\vspace{-5mm}
\end{figure*}

\section{Proposed Method}
This section describes the structure of Horizontal Pyramid Matching(HPM) framework as shown in Fig~\ref{fig:framework}. The input image is fed into the a backbone network to extract the feature maps. After that, we use horizontal spatial pyramid pooling module to obtain spatial information in each local and global spatial bin. For each horizontal spatial bin, we use both global average pooling operation and max pooling operation to obtain the features of global part and most discriminative part of person body. Then, convolutional layers are used to reduce the dimensions of the column feature maps from 2048 to 256 and each column feature is input into a non-share fully connectedly layer and followed by a softmax function to predict the ID of each input image. During testing, all these features are concatenated together to obtain the final Re-ID feature representations. More details will be given in the following.
\subsection{Horizontal Pyramid Matching}

\noindent {\bf Backbone Network} The HPM can take various network architecture like VGG~\cite{simonyan2014very}, Resnet~\cite{he2016deep} and Google Inception~\cite{szegedy2016rethinking} as the backbone. Our paper choose the Resnet50 as backbone network with some modifications following the previous state-of-the-art~\cite{sun2017beyond}. First, the average pooling layer and the fully connected layer are removed. Also, the stride of the conv4\_{1} is set to 1. As a result, the size extracted feature maps will be $\frac{1}{16}$ of the input image size.

\noindent {\bf Horizontal Pyramid Pooling (HPP) module}
HPP is inspired by Special Pyramid Pooling (SPP)~\cite{he2014spatial}, which is proposed to eliminate uncertain length of feature vectors caused by different input sizes of images. The differences between our HPP module and SPP mainly include two aspects: 1) {\bf motivation: } HPP is designed to learn to enhance the discriminative information of partial person body at various scales, while SPP is to address the issue of inconsistent length of image feature vectors. 2) {\bf operation: } Since the distribution of distinguish partitions of a person is from head to foot, HPP slices the feature maps into multiple scrips in a horizontal manner, which is different from SPP using a 2-D spatial manner. With HPP, we can obtain vectors of fixed length for person parts at different horizontal pyramid scales. These vectors are further fed into one convolutional layer and one fully-connected layer for learning classification. In this way, the discriminative ability of person parts can be captured from global to local, from coarse to fine.    


Formally, denote the feature maps extracted by the backbone network as $F$. We adopt 4 pyramid scales within the HPP module and $F$ is sliced into several spatial bins horizontally and equally according to different scales. Specifically,
assume each spatial bin as $F_{i, j}$. $i, j$ stand for the index of scale and the index of bins in each scale. For instance, $F_{3, 4}$ means the fourth bin in third pooling scale. Then, we pool each spatial bin $F_{i, j}$ by global average and max pooling to generate column feature vector, $G_{i, j}$. $$G_{i, j} = avgpool(F_{i, j}) + maxpool(F_{i, j})$$ After that, each $G_{i, j}$ is fed into a convolutional layer to reduce the dimensions from 2048 to 256, denote as $H_{i, j}$. These $H_{i, j}$ with the same i can be considered as a description of the person. This kind of description covers more detailed partial features with the increasing pyramid scales.

\subsection{Loss Function}
We leverage the classification-based model to tackle the person Re-ID task. Therefore, the target is to predict the ID of each person, thus person-specific feature representations can be then learned by the optimized classification model. We use a branch of fully connected layer as the classifier, each feature column vector $H_{i,j}$ is fed into a corresponding classifier $FC_{i,j}$ and following a softmax function to predict its ID. During training, the output of given image $I$ is a set of predictions $\hat{y}_{i, j}$. Each $\hat{y}_{i, j}$ can be formulated as 
$$\hat{y}_{i, j} = \argmax_{c\in P} \frac{exp((W_{i, j}^{c})^{T} H_{i, j}(I))}{\sum_{p=1}^{P}exp((W_{i, j}^{p})^{T} H_{i, j}(I))}$$
where the P is the total number of person ID, $W_{i,j}$ is the learnt weights of $FC_{i,j}$, $y$ is the ground truth ID of input image $I$. The loss function is sum of Cross Entropy loss of each output $\hat{y}_{i,j}$.
$$Loss = \sum_{n=1}^{N}\sum_{i,j} CE(\hat{y}_{i, j}^{n}, y^{n})$$
where N is the size of mini-batch, CE is the Cross Entropy loss.

\subsection{Variant of HPM}
HPM may have some variants different from the basic framework describe above, ~\eg different pyramid scales and pooling strategies. 

{\bf Number of pyramid scales}
The HPM can have several different number of scales. Instead of the 4 scales, it can be any number up to the $log_{2}(h)$, where $h$ is the height of feature map. The HPM structure with different pyramid scales is shown in Table~\ref{method:t1}.The model will focus on more detailed and fine partitions of the given person with the increasing of pyramid scales. Since our loss function is a linear combination of each pyramid scales, if there are too many pyramid scales, the global information of the person may be underestimated. On the other hand, if too few pyramid scales, the features of local discriminative partition may be more difficult to extract. Thus, choosing a proper pyramid scales that can balance the global and local features is vital for the performance of HPM. 
\begin{table}[ht]\setlength{\tabcolsep}{5pt}
\centering
\begin{tabular}{c|l|l}
\hline
\# Pyramid Scale & \# Spatial Bins & Size of Spatial Bins \\ \hline
1 & 1 & 24x8\\ 
2 & 1, 2 & 24x8, 12 x 8\\
3 & 1, 2, 4 & 24 x 8, 12x8, 6x8 \\
4 & 1, 2, 4, 8 & 24 x 8, 12x8, 6x8, 3x8 \\
\hline
\end{tabular}
\caption{HPM Structure with different pyramid scales.}
\label{method:t1}
\end{table}

{\bf Pooling strategies}
The HPM uses both average pooling and max pooling. The global average pooling is a traditional operation in many classification framework, because it enforces a corresponding relation between feature maps and categories. However, the global average pooling can lose some very discriminative information by the average operation. For example, if one partition of the person is very discriminative but surrounded by background, in this case, the global average pooling will obtain the average of the discriminative part and the background region, which may lead to a low response and miss it. To deal with this problem, we use both average pooling and max pooling, which can maintain the global relation with the identification and preserve the discriminative part.

We will provide extreme ablation experiments in the following section to validate the effectiveness of our settings.
\section{Experiments}
\subsection{Dataset and Evaluation Protocol}
{\bf Market1501}~\cite{zheng2015scalable} contains 32,668 images of 1,501 labeled persons of six camera views. There are 19,732 gallery images and 12,936 training images detected by DPM~\cite{felzenszwalb2010object}, including 751 identities in the training set and 750 identities in the testing set. It also contains 500,000 images as some distractors, which may has a considerable influence on the retrieval accuracy.

{\bf DukeMTMC-ReID}~\cite{ristani2016performance,zheng2017unlabeled} is a subset of the DukeMTMC dataset. It contains 1,812 identities captured by 8 cameras. There are 2,228 query images, 16,522 training images and 17,661 gallery images, with 1,404 identities appear in more than two cameras.Also, similar with the Market1501, the rest 408 IDs are considered as distractor images. DukeMTMC-ReID is one of the most challenging re-ID datasets up to now with so many images from 8 multi-cameras.

{\bf CUHK03}~\cite{li2014deepreid} consists of 14,097 cropped images from 1,467 identities. For each identity, images are captured from two cameras and there are about 5 images for each view. There are two ways to obtain the annotations: human labeled and detected by DPM. Our evaluation is based on the detected label image.

{\bf Evaluation Protocol}
In our experiment, we use Cumulative Matching Characteristic (CMC) curve and the mean average precision (mAP) to evaluate our approach. CMC represents the accuracy of the person retrieval, it is accurate when each query only has one ground truth. However, when multiple ground truths exist in the gallery, the goal is to return all right match to user. In this case, CMC may not have enough discriminative ability, but the mAP could reflect the recall. For Market-1501 and DukeMTMC-ReID. We use the evaluation packages provided by~\cite{zheng2015scalable} and~\cite{zheng2017unlabeled}, respectively. And for CUHK03, we adopt the new training/testing protocol proposed in~\cite{zhong2017re}. Moreover, for simplicity, all results reported in this paper is under the single-query setting and does not use the re-ranking proposed in~\cite{zhong2017re}. 
\subsection{Implementation Details}
In order to obtain enough information from person image and proper size of feature map for horizontal pyramid pooling, we resize all the image to 384x128. For the backbone network, we use Resnet50 that initialized with the weights pretrained on ImageNet~\cite{deng2009imagenet}. We remove the last fully connected layer and average pooling layer and set the stride of last resent conv4\_{1} from 2 to 1. The training images are augmented with horizontal flipping and normalization. The batch size is set to 64 and we train model for 60 epoch. The base learning rate is set to 0.1 and decay to 0.01 after 40 epochs. Notice that learning rate for all pretrained Resnet layer is set to 0.1 x base learning rate. The stochastic gradient descent (SGD) with 0.9 momentum is implemented in each mini-batch to update the parameters. During evaluation, we concatenate the feature vectors after the $1x1$ convolution operation together to generate feature representation of query image.  The feature from original image and horizontal flipped image are added up and normalized feature for retrieval evaluation. Our model is implemented on Pytorch platform and train with two NVIDIA TITAN X GPUs. All datasets share the same experiments setting as above. 

\begin{figure}
	\centering
	\includegraphics[width=0.45\textwidth]{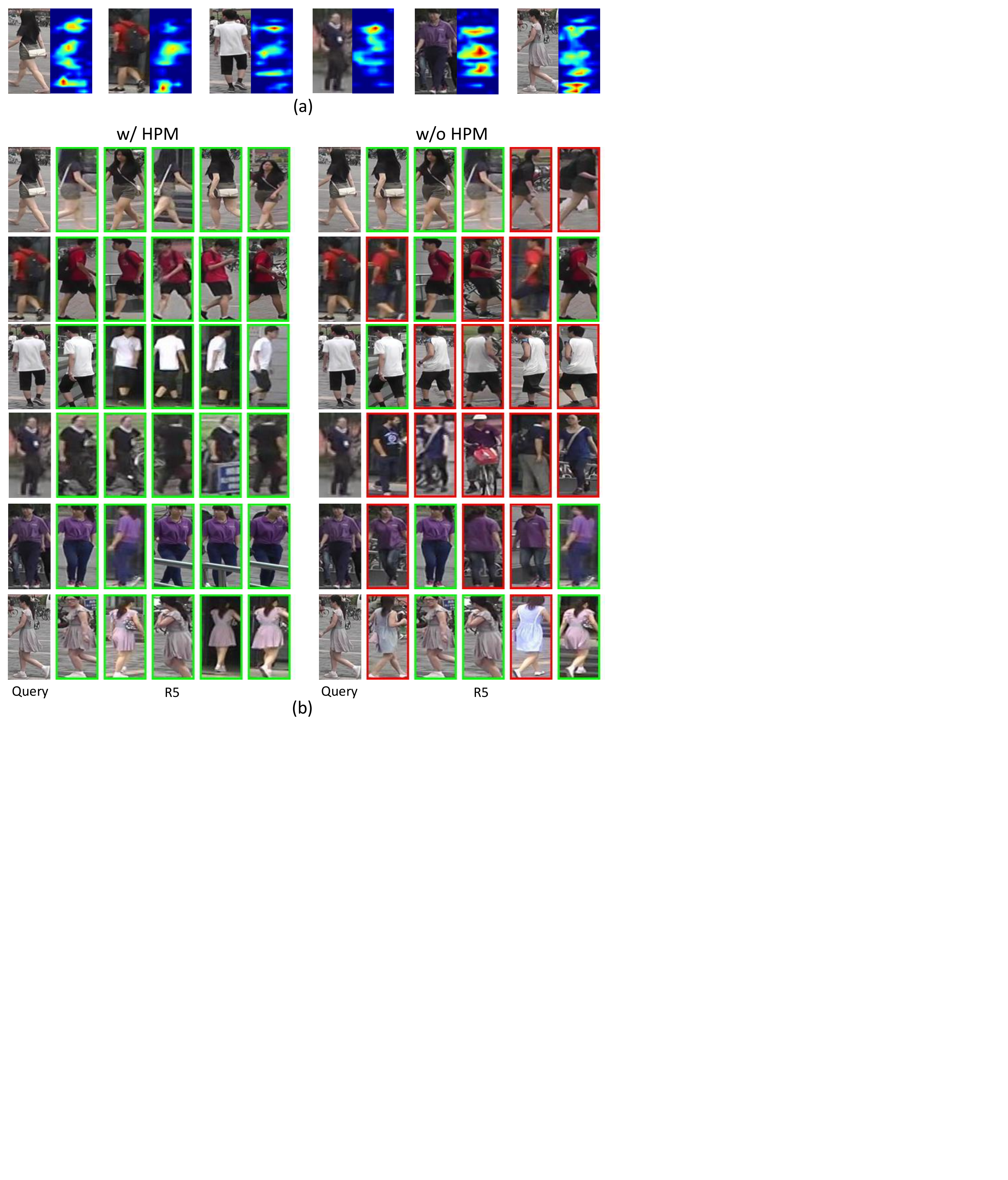}
	\caption{Qualitative Results: (a) Queries and the corresponding discriminative heatmaps learned by the proposed HPM. (b) Comparisons of R5 of w/ HPM and w/o HPM schemes.}
	\label{fig:samples}
	\vspace{-5mm}
\end{figure}

\begin{table*}[t]\setlength{\tabcolsep}{8pt}
\centering
\begin{tabular}{l|c|c|c|c|c|c|c|c}
\hline
\multirow{2}{*}{Model} & \multicolumn{4}{c}{Market1501}  & \multicolumn{2}{|c}{DukeMTMC-ReID} & \multicolumn{2}{|c}{CUHK03}  \\ \cline{2-9}
& R1 & R5 & R10 & mAP & R1 & mAP & R1 & mAP \\ \hline
BoW+kissme \cite{zheng2015scalable} & 44.4 & 63.9 & 72.2 & 20.8 & 25.1 & 12.2 & 6.4 & 6.4\\
WARCA \cite{jose2016scalable} &  45.2 & 68.1 & 76.0 & -- & -- & -- & -- & --\\
SVDNet \cite{sun2017svdnet} &  82.3 & 92.3 & 95.2 & 62.1 & 76.7 & 56.8 & 41.5 & 37.3\\ 
PAN \cite{zheng2017pedestrian} &  82.8 & -- & -- & 63.4 & 71.6 & 51.5 & 36.3 & 34.0\\
PAR \cite{Zhao2017DeeplyLearnedPR} &  81.0 & 92.0 & 94.7 & 63.4 & -- & -- & -- & --\\
MultiLoss \cite{Li2017PersonRB} &  83.9 & -- & -- & 64.4 & -- & -- & -- & --\\
TripletLoss \cite{hermans2017defense} &  84.9 & 94.2 & -- &69.1 & -- & -- & --\\
MultiScale \cite{Chen2017PersonRB} & 88.9 & -- & -- & 73.1 & 79.2 & 60.6 & 40.7 & 37.0 \\
MLFN \cite{chang2018multi} & 90.0 & -- & -- &  74.3 & 81.0 &  62.8 & 54.7 & 49.2\\ 
HA-CNN \cite{li2018harmonious} &  91.2 & -- & -- & 75.7 & 80.5 & 63.8 & 41.7 & 38.6\\
AlignedReID \cite{zhang2017alignedreid} & 91.0& 96.3& -- & 79.4 & -- & -- & -- & -- \\
Deep-Person \cite{bai2017deep} & 92.3 & -- & -- & 79.5 & 80.9 & 64.8 & -- & --\\
PCB \cite{sun2017beyond} & 92.4 & 97.0 & 97.9 & 77.3 & 81.8 & 66.1 & 61.3 & 54.2\\
PCB + RPP\cite{sun2017beyond} &  93.8 & 97.5 & 98.5 & 81.6 & 83.3 & 69.2 & 63.7 & 57.5 \\ \hline
HPM(ours) & {\bf 94.2} & 97.5 & {\bf 98.5} &  {\bf 82.7} & {\bf 86.6} & {\bf 74.3} & {\bf 63.9} & 57.5 \\ \hline
\end{tabular}
\caption{Comparison of the proposed method with the state-of-art on Market-1501, DukeMTMC-ReID and CUHK03 with new protocol. HPM is implemented with four pyramid scales and combine both average pooling and max pooling described in Fig~\ref{fig:framework}.}
\vspace{-2mm}
\label{exp:soa}
\end{table*}

\begin{table}\setlength{\tabcolsep}{2pt}
\centering
\begin{tabular}{l|c|c|c}
\hline
\multirow{2}{*}{Model}  &  \multicolumn{3}{c}{Market1501}  \\ \cline{2-4}
& R1 & R5 & mAP  \\ \hline
No Pyramid Structure & 88.1 & 94.6 & 71.2 \\ \hline
HPM  & {\bf 94.2}(+6.1) & {\bf 97.5}(+2.9) & {\bf 82.7}(+11.5) \\ \hline 
\end{tabular}
\caption{Evaluation of effectiveness of Pyramid Structure, HPM uses four pyramid scales and combine both average pooling and max pooling, Non pyramid structure split the feature into same partitions as the last scale in HPM but without pyramid structure}
\label{exp:ab2}
\end{table}

\begin{table*}[t]\setlength{\tabcolsep}{3pt}
\centering
\begin{tabular}{l|c|c|c|c|c|c|c|c|c|c|c|c|c}
\hline
\multirow{2}{*}{Model} & \multirow{2}{*}{Feature Dim} &  \multicolumn{4}{c}{Market1501}  & \multicolumn{4}{|c}{DukeMTMC-ReID} & \multicolumn{4}{|c}{CUHK03}  \\ \cline{3-14}
& & R1 & R5 & R10 & mAP & R1 & R5 &R10 & mAP & R1 & R5 & R10 & mAP \\ \hline
HPM + \#PS 1 + Avg pool & 256 &88.1 & 94.6 & 96.4 & 71.2 & 79.3 & 89.7 & 91.9  & 61.0 & 39.2 & 61.1 & 71.6 & 37.3 \\ \hline
HPM + \#PS 2 + Avg pool & 256x(1+2) & 92.0 & 96.9 & 97.9 & 78.3 & 83.1 & 91.9 & 93.4 & 68.9 & 53.2 & 73.2 & 79.6 & 48.9 \\ \hline
HPM + \#PS 3 + Avg pool & 256x(1+2+4) &92.3 & 97.2 & 97.9 & 79.3 & 84.5 & 92.4 & 94.1 & 70.8& 58.2 & 76.7 &  83.1 & 52.8 \\ \hline
HPM + \#PS 4 + Avg pool &256x(1+2+4+8)& 93.2 & 97.3 & 98.1 & 79.5 & 84.8 & 92.5 & 94.1 & 72.1 & 58.6 & 76.8 & 83.8 & 53.4 \\ \hline
HPM + \#PS 4 + Max pool &256x(1+2+4+8)& 93.6 & {\bf 97.7} & 98.3 & 81.6 & 86.2 & {\bf 93.2} & 94.8 & 74.1 & 62.4 & 78.9 & {\bf 86.3} & 57.4 \\ \hline
HPM + \#PS 4 + Max+Avg pool & 256x(1+2+4+8) & {\bf 94.2} & 97.5 & {\bf 98.5} & {\bf 82.7} & {\bf 86.6} & 93.0 & {\bf 95.1} & {\bf 74.3} & {\bf 63.9} & {\bf 79.7} & 86.1 & {\bf 57.5} \\ \hline
\end{tabular}
\caption{Performance comparison of the proposed method with different pyramid scales and different pooling strategies as described in Section3.4. PS is the abbreviation of Pyramid Scales.}
\vspace{-5mm}
\label{exp:ab1}
\end{table*}


\subsection{Comparison with the State-of-the-arts}
{\bf Results on Market1501} Comparisons between HPM and state-of-art approaches on Market1501 are shown in Table~\ref{exp:soa}. The results show that our HPM achieves the mAP of 83.1\% and Rank 1 accuracy 94.2\%, which both surpass all existing works more than 1.5\% and 0.4\%, respectively. It should be noted that we do not conduct any post-processing operation (~\eg the re-rank algorithm given by~\cite{zhong2017re}), which can further bring a considerably improvement in terms of mAP. PCB~\cite{sun2017beyond} is closest competitor, which also leverages partial-based leaning for person Re-ID. However, there are mainly two disadvantages of PCB,~\ie 1) it splits the features maps into pre-defined patches (6 in PCB) with the assumption that most persons in the given images are well aligned, which not make scene and resist to some outliers; 2) its state-of-the-art results are benefited from a powerful post-processing approach called RPP, which enable the optimized model cannot be trained in an end-to-end manner. In contrast, our HPM splits the feature maps according to various pyramid scales, which is more robust compared with PCB in addressing the outliers that are not well aligned. In addition, our HPM can be end-to-end learned and we believe that any post-processing operation can make a continued improvements upon the current results. From Table~\ref{exp:soa}, we can observe that our HPM makes a 5.4\% improvements compared with PCB in mAP. Even without post-processing, our HPM is still better than PCB+RPP (~\ie 83.1\% \emph{vs.} 81.6\%). Beyond PCB, the best model aims to deal with different size of person MultiScale~\cite{chen2017multi} yields the mAP of 73.1\% and Rank 1 accuracy of 88.9\%. Our HPM model outperforms it by 5.3\% and 10.0\% on Rank 1 and mAP, respectively.

{\bf Results on DukeMTMC-ReID} Person Re-ID results on Duke MTMC-ReID are given in Table~\ref{exp:soa}. This dataset is challenging because it has 8 different camera and the person bounding box size varies drastically across different camera views, however, our HPM achieves even better performance on this dataset. Without any post-processing, it still achieves 74.8\% on mAP and 86.6\% on Rank 1 accuracy, which is better than all other state-of-the-art methods by a large margin, 5.3\% and 3.3\%. Note that our HPM is the first model that can achieve above 80\% on mAP, which surpass all state-of-the-art methods by more than 5.0\%

{\bf Results on CUHK03} Table~\ref{exp:soa} shows results on CUHK03 when detected person bounding boxes are used for both training and testing. HPM achieves the best result of mAP, 59.7\% under this setting. Although the Rank 1 accuracy of HPM is a little lower than PCB+RPP, there's a clear gap, more than 2\%, between HPM and other methods, including the PCB+RPP, on mAP.  And We believe that, as a end-to-end and part-based model, the RPP can also boost the performance HPM further.

{\bf Qualitative Result} We visualize some examples in Figure~\ref{fig:samples}. Concretely, Figure~\ref{fig:samples} (a) shows the queries and the corresponding heatmaps\footnote{We normalize each feature map of the last convolutional feature maps and sum them together to obtain the heatmap.} of the last convolutional feature maps. We observe that the discriminative abilities of multiple person parts are enhanced with our HPM. Figure~\ref{fig:samples} (b) compares the Re-ID results of w/ HPM and w/o HPM schemes. It can be seen that our HPM is very effective in guaranteeing accurate Re-ID results.

\subsection{Ablation Study}
To verify the effectiveness of each component and setting of HPM, we design several ablation study with different settings on Market-1501, DukeMTMC-ReID and CUHK03, including different number of pyramid scales, w/ and w/o using pyramid structure, different pooling strategies. Note that all unrelated settings are the same as HPM implementation detailed in Section 4.2

{\bf Effectiveness of Pyramid Structure}
Previous analysis shows that the HMP reaches the best performance with four pyramid scales, which has up to 8 partial bins on the feature map. In order to verify the effectiveness of pyramid structure, we remove other branches and just preserve the branch with 8 partial bins. From Table~\ref{exp:ab2}, we can observe that Rank 1, Rank 5 and mAP drop from 94.2\%, 97.5\% and 82.7\% to 92.0\%, 96.3\% and 76.4\%, respectively. Such an operation is similar to PCB, which leverage 6 partial bins. The reason is that many persons are not well aligned in the images, and naively split the feature maps into a pre-defined number of bins cannot well resist to unaligned outliers. In addition, discriminative information may hardly be learned for some parts if we apply too dense division. In contrast, with our pyramid structure, we can formulate partial features from coarse to fine, which can finally form into a more robust feature representation for person images.

\begin{figure}
\begin{center}
\bgroup 
 \def\arraystretch{0.2} 
 \setlength\tabcolsep{1.0pt}
\begin{tabular}{cc}
\includegraphics[width=0.5\hsize]{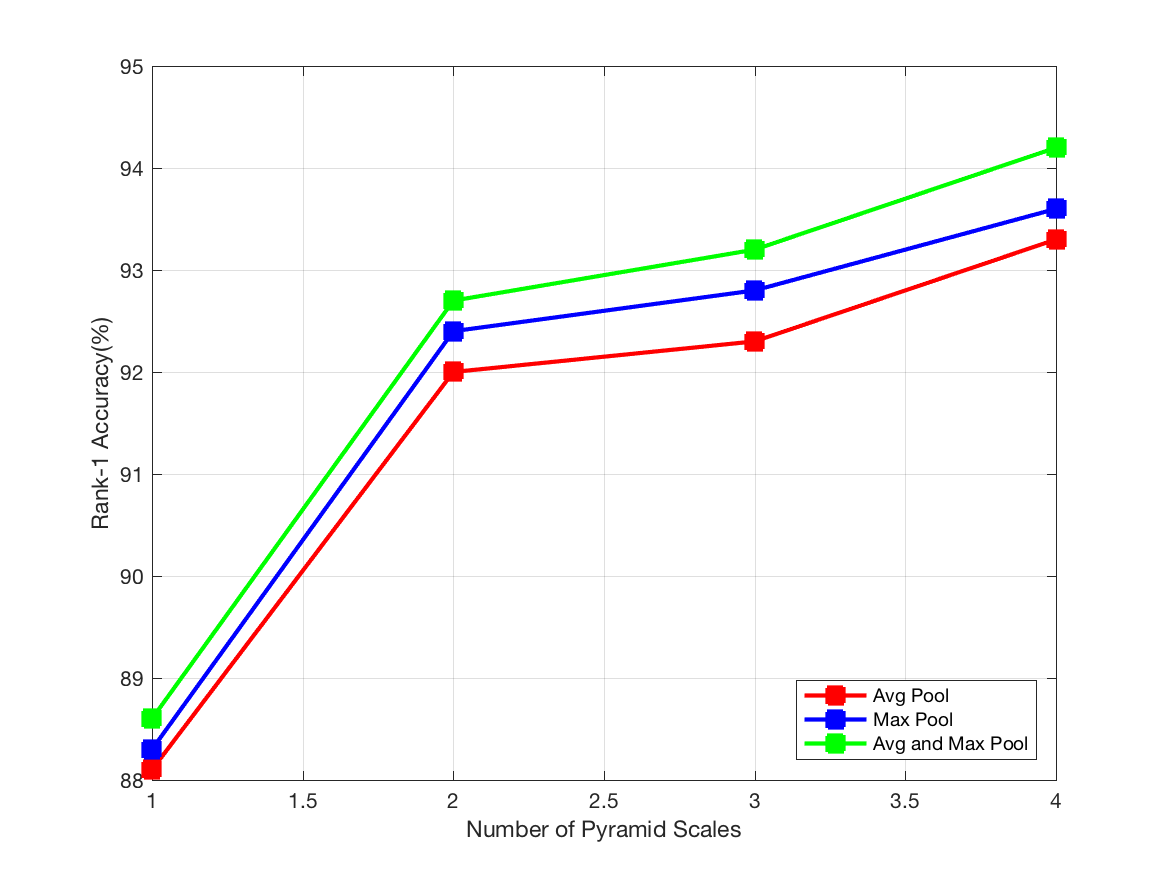} &
\includegraphics[width=0.5\hsize]{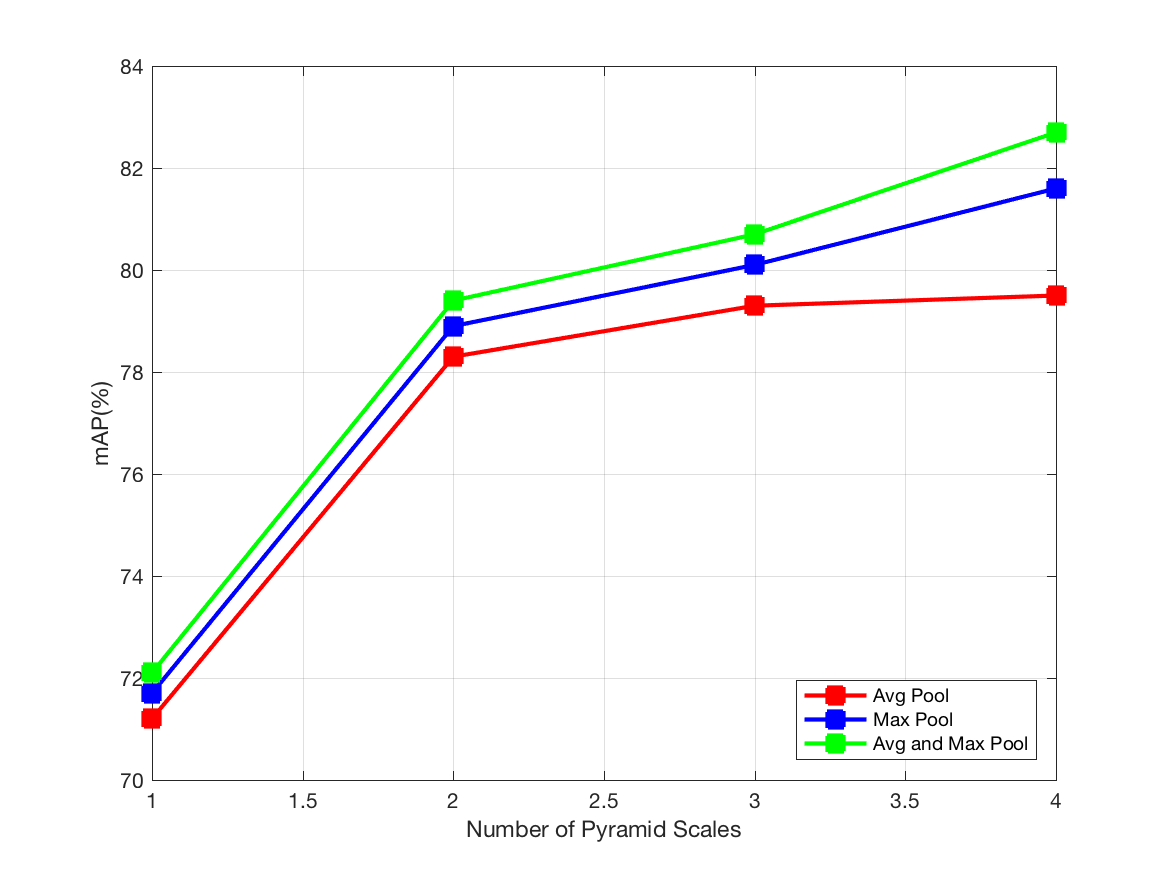}
\end{tabular} \egroup
\end{center}
\caption{Impact of pyramid scales. Rank-1 accuracy and mAP are compared.}
\vspace{-5mm}
\label{exp:compare}
\end{figure}

{\bf Number of Pyramid Scales}
Table~\ref{exp:ab1} shows the Re-ID results of HPM with different pyramid scales, ~\eg 1, 2, 4, 8. From these results, we can find that HPM reaches the best performance with four pyramid scales. Intuitively, the number of pyramid $p$ determines the granularity of the partition feature. When $p=1$, it is equivalent to global pooling. With the increasing the number of $p$, Rank 1 accuracy and mAP are significant improved from 88.1\% and 71.2\% to 93.2\% and 79.5\%, as illustrated in Fig~\ref{exp:compare}. The reason why the HPM does not drops dramatically at some point as introduced in~\cite{sun2017beyond} is that the pyramid structure can combine both global and local features, which may increase the discriminative ability of very small partition. Since the last convolutional feature maps are with 24 horizontal units, we also try more dense pyramid scales, such as 12 and 24. However, more pyramid scales will bring additional computational cost and there is no obvious improvement can be observed. Therefore, we finally adopt 4 pyramid scales in this work.

{\bf Pooling Strategies}
Row4 and Row 5 in Table~\ref{exp:ab1} shows the performance of HPM with different pooling strategies. It can be observed that max pooling performs better than average pooling in the most cases. The reason is that average pooling will take all locations of a particular parts into account and all the locations contribute equally to the final partial representation. Thus, the discriminative ability of the representation produced by average pooling can be easily influenced by the unrelated background patterns. In contrast, the global max pooling only preserve the largest response values for a local view. We consider these two pooling strategies are complementary in producing feature representations from global and local vies. Therefore, we integrate them into a unified model to take advantages from these two strategies. Experimental results in Table~\ref{exp:ab1} demonstrate that mixing the two pooling strategies achieves better results compared with using either of them. 
 

\section{Conclusion}
In this work, we propose a novel Horizontal Pyramid Matching (HPM) approach to address the challenging person Re-ID task. The proposed HPM exploits partial information of each person to Re-ID, which successfully enhances the discriminative ability of partial feature and finally forms the into a more robust feature representation for the target person. In addition, we leverage both partial-based global average and max pooling to mine the discriminative information of each part in a global-local manner, which can further improve the robustness of partial features. All the components detailed in this work can be easily embedded into any other framework to make a further performance improvement. Extensive ablation studies and comparisons well demonstrate the effectiveness of our HPM approach. In the future, we are plan to simultaneously optimize Re-ID and other related tasks, such as human activity recognition~\cite{dai2017temporal,dai2017efficient,dai2018tan}.


\vspace{3mm}
\noindent
\textbf{Acknowledgements.} This work is in part supported by IBM-ILLINOIS Center for Cognitive Computing Systems Research (C3SR) - a research collaboration as part of the IBM AI Horizons Network. Shi is supported in part by IARPA Deep Intermodal Video Analytics (IARPA DIVA). Fu and Zhou are supported by CloudWalk Technology.

\bibliographystyle{aaai}
\bibliography{aaai}
\end{document}